\documentclass[letterpaper, 10 pt, conference]{IEEEtran}  

\IEEEoverridecommandlockouts                              

\usepackage{times}
\usepackage{amsmath}
\usepackage{amsfonts}
\usepackage{graphicx}

\usepackage[numbers]{natbib}
\usepackage{multicol}
\usepackage{multirow}
\usepackage{float}
\usepackage[bookmarks=true]{hyperref}
\usepackage[
singlelinecheck=false 
]{caption}
\usepackage{xcolor}
\usepackage{enumitem}

\usepackage{algorithm}
\usepackage{algorithmic}

\usepackage{placeins}

\makeatletter
\newcommand\fs@norules{\def\@fs@cfont{\bfseries}\let\@fs@capt\floatc@ruled
  \def\@fs@pre{}%
  \def\@fs@post{}%
  \def\@fs@mid{\kern3pt}%
  \let\@fs@iftopcapt\iftrue}
\makeatother
\floatstyle{norules}
\restylefloat{algorithm}

\makeatletter
\patchcmd{\@makecaption}
  {\scshape}
  {}
  {}
  {}
\makeatother
\title{\LARGE \bf
Composing Diffusion Policies for Few-shot Learning of Movement Trajectories
}
\author{Omkar Patil, Anant Sah, Nakul Gopalan
\thanks{School of Computing and Augmented Intelligence, Arizona State University }%
}

\begin{document}

\maketitle
\thispagestyle{empty}
\pagestyle{empty}

\begin{abstract}
    Humans can perform various combinations of physical skills without having to relearn skills from scratch every single time. For example, we can swing a bat when walking without having to re-learn such a policy from scratch by composing the individual skills of walking and bat swinging. Enabling robots to combine or compose skills is essential so they can learn novel skills and tasks faster with fewer real world samples. To this end, we propose a novel compositional approach called DSE- Diffusion Score Equilibrium that enables few-shot learning for novel skills by utilizing a combination of base policy priors. Our method is based on probabilistically composing diffusion policies to better model the few-shot demonstration data-distribution than any individual policy. Our goal here is to learn robot motions few-shot and not necessarily goal oriented trajectories. Unfortunately we lack a general purpose metric to evaluate the error between a skill or motion and the provided demonstrations. Hence, we propose a probabilistic measure – Maximum Mean Discrepancy on the Forward Kinematics Kernel (MMD-FK), that is task and action space agnostic. By using our few-shot learning approach DSE, we show that we are able to achieve a reduction of over 30\% in MMD-FK across skills and number of demonstrations. Moreover, we show the utility of our approach through real world experiments by teaching novel trajectories to a robot in 5 demonstrations.
\end{abstract}


\section{Introduction}
\begin{figure*}
\centering
\includegraphics[width=0.8\linewidth]{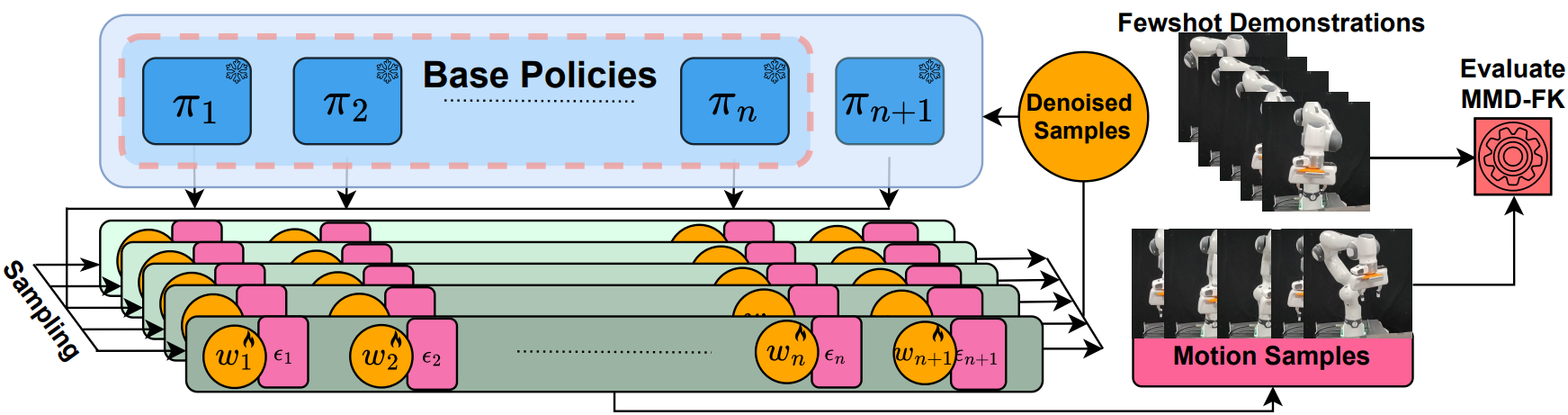}
\caption{An outline of our approach. We assume a set of base policies $\pi_i$, $i=1..N$ and train another policy $\pi_{N+1}$ on the provided demonstrations. We compose over these policies and infer the compositional weights using an optimization procedure with objective of Maximum Mean Discrepancy on the Forward Kinematics Kernel (MMD-FK). Only one optimization cycle is shown in the image.}
\vspace{-11pt}

\label{img:meth}
\end{figure*}
For robots to be deployed in unstructured environments and interact with humans, they should be capable of learning new skills from very few demonstrations. 
For example, wiggling the end-effector while moving forward to clean a table is a combination of two independent motions.
This wiggling motion can be combined with different primitive motions to clean floors, to wash dishes, to fit a bed-sheet, to iron a cloth, etc. 
These are not goal oriented trajectories, but continuous motions that are sometimes dynamical trajectories in configuration space where a robot follows a sequence of movements.
Robots should not be expected to learn these composed motions one at time but rather combine previously learned skills along with utilizing any given demonstrations. However, finding the right skills to combine from a base set and the extent of their contributions in the resulting motion is non-trivial. Existing compositionality methods either directly pick and choose the priors to compose while only learning the ratios of the priors' contribution~\cite{peng2019mcp}, or do not have a method 
to utilize residual information in the provided demonstrations~\cite{urain2023composable,wang2024poco}.

To tackle these shortcomings, we propose Diffusion Score Equilibrium(DSE), a compositional method that works over a set of base policies by inferring the extent of their contribution given a few  demonstrations. Importantly, our method does not assume the policies to compose for achieving the desired behavior, and scales the contribution of base policies based on the information available in the provided demonstrations. A core element of our approach is inferring the contribution of each base policy in the resulting behavior, which we refer to as compositional weights henceforth. We infer these weights by minimizing the distance between a proposed trajectory and the few-shot demonstration data-distribution.  

Underlying our approach is the insight that composing diffusion models can result in novel motion generation that interpolates between the individual distributions. We leverage this insight to efficiently learn a novel skill by interpolating between the noisy distribution learned from the few demonstrations of a novel skill and the set of base policy distributions for minimizing the distance to the few-shot demonstration data-distribution.

However, an impediment to this approach is the lack of a general purpose metric that can measure this distance between a set of trajectories. Most metrics in robotics and within compositional works use task specific metrics such as success rate~\cite{wang2024poco, urain2023composable} that are specific to their chosen task and actions. This however does not measure the quality of the composed distribution with respect to the reference trajectories or even base policies. To fill this gap and evaluate the compositional weights in our method, we propose Maximum
Mean Discrepancy on the Forward Kinematics Kernel (MMD-FK) based on Maximum Mean Discrepancy~\cite{gretton2012kernel} and the Forward Kinematics kernel \cite{das2020forward}. Our metric measures the distance between two robot trajectory distributions considering \emph{all the physical links in the robot's body and not just the end effector.}

We show that by inferring the compositional weights by minimizing MMD-FK, our method DSE scales with the number of provided demonstrations and achieves superior performance in both low and high data regimes. DSE results in $30\%$ to $50\%$ lower MMD-FK error in different data regimes than a demonstration fine-tuned policy and is also superior to prior compositional approach using diffusion models. 
Our contributions in this work are as follows-
\begin{itemize}[leftmargin=5mm, nolistsep]
    \item We present a novel compositional approach for sample-efficient learning called Diffusion Score Equilibrium (DSE). Our method does not rely on manually choosing which base policies to compose, and scales the performance with the number of demonstrations provided for the new skill. To the best of our knowledge, our work is also the first to learn compositional weights over a set of diffusion policies from the target demonstrations.
    \item We propose MMD-FK to fill the gap of a task and action space agnostic metric. We use the novel combination of the distributional MMD measure with the Forward Kinematics kernel to calculate distances between two trajectory distributions over the whole body of the robot.
    \item We showcase that our approach is superior to simple probabilistic composition of base policies and even training a model on the demonstration data. We showcase our results on nine non-orthogonal base policy priors and with multi-modal priors for several new trajectories that the robot has not learned before. Moreover, our real world experiments to teach the robot novel trajectories from a few demonstrations showcase the robustness and utility of DSE with noisy real robot data to learn policy compositions.
\end{itemize}

\section{Background}
\label{sec:backg}
\subsection{Related work demonstrating compositionality in learning}

Large scale robot learning has shown considerable promise in terms of generalization. However, it requires large computational overheads and massive data collection, which is challenging for robotics. Efficiency of learning new skills and being able to reuse them will be crucial for robot learning to be successful in unstructured environments. Compositionality is one such approach that aims to maximize the reuse of policies or representations. Compositionality can be induced in various aspects of the policy learning process for sample efficiency gains. Functional \cite{mendez2023reuse}, representational \cite{andreas2019measuring}, temporal \cite{bacon2016optioncritic,wan2024lotus} and policy composition \cite{hinton2002training,peng2019mcp,wang2024poco} feature prominently in the machine learning and robotics literature. Policy composition enables embodied agents to reuse two or more of their learned policies together for sampling from a novel distribution. Previously, a mixture of experts~\cite{jacobs1991adaptive} such as a weighted average or a product of distributions~\cite{hinton2002training} representing sub-policies has been used to model skill, motion or constraint composition. Unlike other forms, in policy composition, the distributions learned for motions or constraints are probabilistically composed and the resulting sample is generated from the new probability landscape.

In supervised learning, traditionally energy based models have been utilized to probabilistically compose distributions \cite{du2020compositional}, since the energy functions could be added to obtain a product of corresponding distributions, from which samples could then be generated using Markov Chain Monte Carlo (MCMC). Recently, the number of works leveraging compositionality has surged owing to the discovery of the connection between energy based models and diffusion models and the ease of training the latter \cite{du2023reduce}. \citet{liu2024composable} and \citet{yang2023compositional} compose learned constraints for generalizing to their novel combinations, similar to the work of \citet{liu2021learning} on composing learned object relations in images. Relevant to our work, \citet{urain2023composable} use energy based models to compose different motions for obstacle avoidance. However, the distributions are handcrafted and not learned. \citet{paraschos2013probabilistic} use Bayesian hierarchical models which are challenging to adapt to novel domains and require significant domain knowledge to tune. Our approach is more intuitive to train without requiring as much domain knowledge. \citet{wang2024poco} compose learned distributions for the task of robotic tool use, but these policies are task-conditional and unconditional versions of the same task, and do not result in generation of novel motion.

\vspace{-0.04cm}
\subsection{Diffusion Models}
Our aim is to learn the action distribution $a_0^L$ for a fixed trajectory length $L$ from $D$ demonstrations. Here, we use $a$ to denote action for all the trajectory time-steps for brevity and drop the $L$ notation. Gaussian diffusion models \cite{sohldickstein2015deep} learn the reverse diffusion kernel $p_\theta(a_t|a_{t-1})$ for a fixed forward kernel that adds Gaussian noise at each step $q(a_t|a_{t-1})= \mathcal{N}(a_t; \sqrt{\alpha_t}a_{t-1}, (1-\alpha_t)\mathcal{I})$, such that $q(a_T)\approx\mathcal{N}(0, \mathcal{I})$. Here $t<=T$ represents the diffusion time-step and $\alpha_t$ the noise schedule. To generate trajectories from the learned data distribution $p_\theta(a_0)$, we sample at time step $T$ from $\mathcal{N}(0, \mathcal{I})$ and apply the reverse diffusion kernel $p_\theta(a_t|a_{t-1})$ at each time step. For training the model, maximization of the log-likelihood of the data distribution $log~q(a_0)$ and reparametrization of the forward diffusion kernel yields the following loss used in practise \cite{ho2020denoising}:
\begin{equation}
    \label{eq:diffusion}
    \mathcal{L}_t(\theta)=\mathbb{E}_{q(a_0)\mathcal{N}(\epsilon_0;0,\mathcal{I})}\left[{\lambda_t}[||\epsilon_0 -\hat{\epsilon}_\theta(a_t, o, t)||_2^2]\right]
\end{equation}
Here $\lambda_t$ is a function of $\alpha_t$, and the network $\epsilon_\theta$ is conditioned on observation $o$. We train our model to predict the noise $\epsilon_0$ added to action $a_0$ for generating the noisy action $a_t$ taken as the input to the network. Tweedie's formula \cite{efron2011tweedie} can be used to show that $\epsilon_0$, and consequently $\epsilon_\theta$ are proportional to the score of the diffused data distribution $q(a_t) = \int q(a_t|a_0)q(a_0)da_0$ \cite{luo2022understanding}.
\begin{equation}
    \frac{-1}{\sqrt{1-\Bar{\alpha_t}}}\hat{\epsilon}_\theta(a_t, t) \approx \frac{-1}{\sqrt{1-\Bar{\alpha_t}}}\epsilon_0 =\nabla_a log  \ q(a_t)
\end{equation}

\subsection{Policy Composition and Sampling}
\label{bg:pc}
To sample from a product of distributions, we need the score of the composition at each noise scale of the ancestral sampling chain. Our product distribution can be expressed as $p^{comp}(a_0)=p^1_\theta(a_0)*p^2_\theta(a_0)$, where $a_0$ has been specifically written to reflect that the distributions are composed in the data space. Then the score of the composed distribution at diffusion time $t$ can be written as:
\begin{equation}
    \label{eq:comp-score}
    \nabla_{a_t}log~q^{comp}(a_t)=\nabla_{a_t}log\left(\int\left[\prod q^i(a_0)\right]q(a_t|a_0)da_0 \right)
\end{equation}
A long line of works instead add the individual scores of the distributions being composed $\sum_i\left(\nabla_{a_t}log\left[\int q^i(a_0)q(a_t|a_0)da_0 \right]\right)$, since Equation \ref{eq:comp-score} is not tractable. \citet{du2023reduce} bring this out as the reason for inferior quality of samples from composed image distributions and suggest Annealed MCMC samplers instead of ancestral sampling that does not result in the correct sequence of marginals expected by the reverse diffusion process. However, we utilize this sequence of marginals to interpolate between distributions.


\section{Methodology}

\label{sec:method}
\begin{figure}[ht] 
\centering
\minipage{0.50\columnwidth} 
  \includegraphics[width=\linewidth]{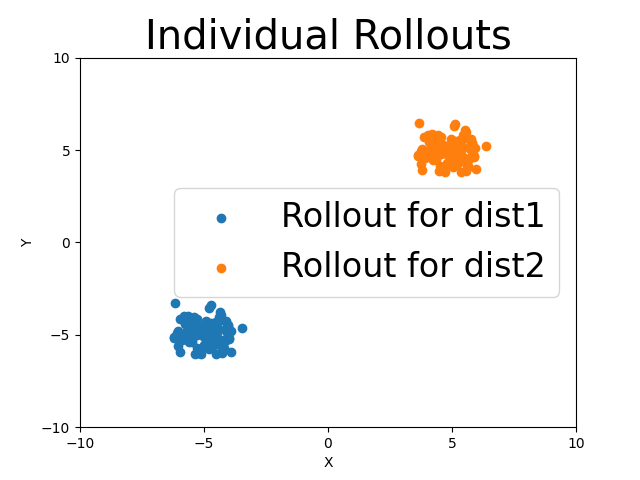}
\endminipage\hfill
\minipage{0.50\columnwidth} 
  \includegraphics[width=\linewidth]{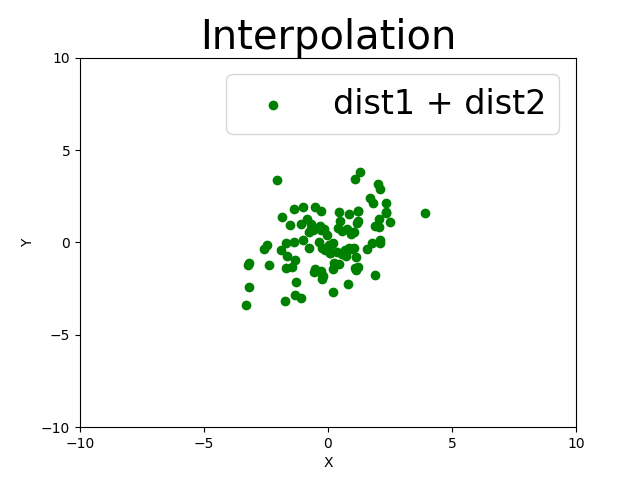}
\endminipage\hfill


\minipage{0.50\columnwidth} 
  \includegraphics[width=\linewidth]{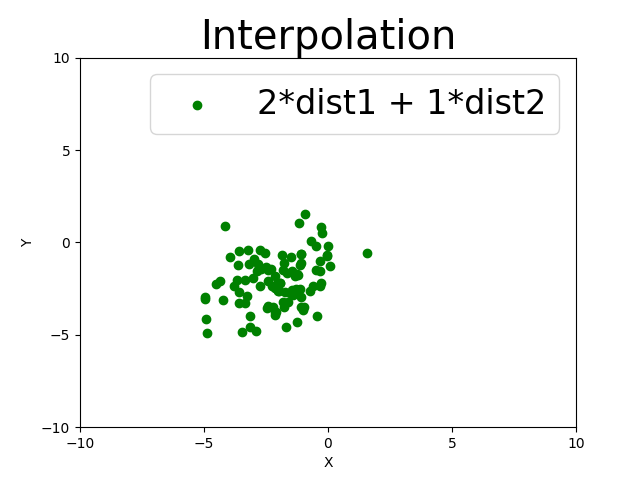}
\endminipage\hfill
\minipage{0.50\columnwidth} 
  \includegraphics[width=\linewidth]{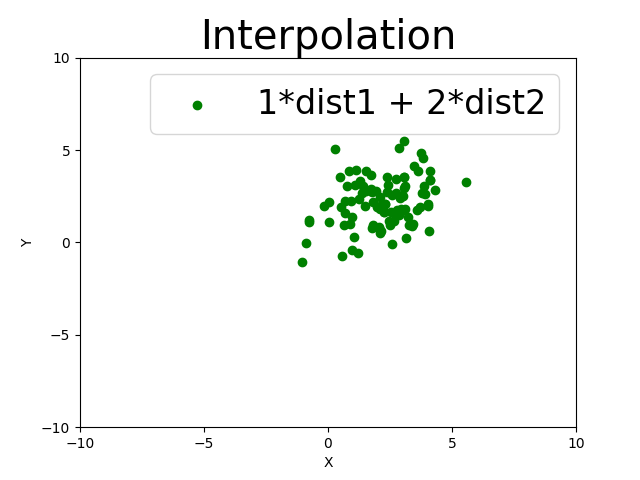}
\endminipage\hfill
\caption{This panel of figures shows the result of 100 samples from Top-left: the individual distributions; Top-right: Derived from the approximated score giving equal weighting to both; Bottom-left: Giving 2:1 weighting to the distribution 1 and 2 respectively; Bottom-right: Giving 1:2 weighting to the distributions. The scores are normalized with respect to their weights. These images clearly show that diffusion models can have an interpolating effect between two distributions based on the weighting of the scores.}
\label{img:res}
\end{figure}

\subsection{Novel Motion Generation by Composing Diffusion Models}
\label{meth:nmg}

To spatially blend between distributions for generating novel motion, we propose to sample from $q^{comp}(a_0)=\prod_{i=1}^{N} q_i(a_0)^{w_i}$, where $\sum_{i=1}^{N} w_i=1$, where we have $N$ base policies. The sum of scores of the composed distribution at each time-step can then be written as:
\begin{scriptsize}
    \begin{equation}
        \label{eq:meth1}
        \nabla_{a_t}log~q^{comp}(a_t) \approx \sum_i^{N}w_i\left(\nabla_{a_t}log~\left[\int q^i(\frac{a_0'}{\sqrt{\Bar{\alpha_t}}})\Phi\left(\frac{a_t-a_0'}{1-\Bar{\alpha_t}}\right)da_0'\right] \right) 
    \end{equation}
\end{scriptsize}
Where $\Phi$ is the standard normal distribution. Here, we have split the mean and variance effects of the forward diffusion transition kernel $q(a_t|a_0)$ to suggest that the individual distributions being composed are not invariant across time-steps.

Expressing the $i^{th}$ base policy distribution at diffusion time-step $t$ as an EBM $p_{i;t}(a)=exp(-E_{i;t}(a))/Z_{\theta}$, we get its score as $\nabla log p_{i;t}(a)=-\nabla E_{i;t}(a)$, where $E_{i;t}$ represents the noisy shifted energy function. Consequently, samples generating using Equation~\ref{eq:meth1} converge to the $exp(-\sum_{i=1}^{N} w_i E_{i,t}(a))/Z_\theta$. This can be interpreted a weighted mixture of noisy energy functions that become more accurate as the diffusion time-step tends to $0$. Changing the weights will relocate the minimums of the composed distribution, with the $w_i=1$ falling back on the base policy $i$. The gradient of the energy function $\nabla E_{i;t}(a)$ is proportional to the output of diffusion models $\hat{\epsilon}_{i;\theta}(a_t, t)$, both of which estimate the score of the data distribution corresponding to the $i^{th}$ base policy \cite{du2023reduce}. Thus a weighted addition of the diffusion model outputs $\sum_{i=1}^{N}{w_i\hat{\epsilon}_{i;\theta}(a_t, t)}$ where $\sum_{i=1}^{N}{w_i}=1$ is proportional to the gradient of the weighted energy function $\nabla \left(\sum_{i=1}^{N}w_iE_{i;t}(a)\right)$ at diffusion time-step $t$. Hence, this enables sampling from regions that are not minimums in any of the individual energy functions or distributions being composed, while also lending some control over it's placement. We would like to emphasize that the denoising process in diffusion models is essential for this as it provides gradual gradient estimates to move towards the target minimums.

To verify our claims, we train two diffusion models on 2D data samples from  Gaussian's with means at $(5,5)$ and $(-5,-5)$ and variance 1. The data distribution has limited overlap. The results on composing them using reverse diffusion sampling are shown in figure \ref{img:res}. On adjusting the weight applied to the score of each model, the samples derived from the approximated score of the composition shift towards the respective distributions, implying that we can blend them in relative proportions of their scores.

\subsection{MMD-FK Metric}
Metrics commonly used in the robotics community are task and action space dependent.  \citet{wang2024poco} and \citet{urain2023composable} use success rate as a metric to evaluate policy composition. However, success rate is not indicative of the quality of the composed trajectory. For evaluating the composition of a set of base policies, we would like to measure the distance between the composed distribution and a reference distribution of trajectories, if available. Several integral probability metrics have been proposed in the image generation literature such a FID \cite{heusel2017gans} and Maximum Mean Discrepancy (MMD) \cite{gretton2012kernel} to quantitatively evaluate the generated samples with respect to the data distribution. Moreover, we would like our metric to measure the distance in the task space where the effect of motion composition is apparent, and not be limited to the end-effector actions. With these requirements in consideration, we propose MMD-FK, a metric that uses the MMD distance on the FK kernel to evaluate the distance between two robot-link trajectory distributions. Our metric for $m$ and $n$ samples from the two distributions respectively can be expressed as: 
\vspace{-6pt}
\begin{equation}
    \begin{aligned}
        \Hat{dist}_{MMD-FK}^2(X, Y) = \frac{1}{m(m-1)}\sum_{i=1}^m\sum_{j\neq i}^m K_{FK}(x_i, x_j) +\\
        \frac{1}{n(n-1)}\sum_{i=1}^n\sum_{j\neq i}^n K_{FK}(y_i, y_j)
        - \frac{2}{mn}\sum_{i=1}^m\sum_{j=1}^n K_{FK}(x_i, y_j) 
    \end{aligned}
\end{equation}

It leverages MMD for it's kernel support that enables measurement of the distance between two distributions in terms of the distance between their feature means in a latent space. To evaluate task-space distances even with action space as the robot configuration, we use the positive-definite Forward Kinematics kernel as suggested in \citet{das2020forward}. Here $K_{FK}$ is the positive-definite Forward Kinematics kernel in Equation \ref{eq:kernel}. Equation \ref{eq:kernel} sums over the $m$ control points defined on the robot, typically associated with each link in the kinematic chain. For our purposes, one control point on each kinematic chain allows us to capture the movements of the links of the robot in the task-space. In Equation \ref{eq:kernel}, $K_{RQ}$ is a second-order rational quadratic kernel $K_{RQ}(x, x')=\left(1+\frac{\gamma}{2}||x-x'||^2\right)^{-2}$, with the width of the kernel being $\gamma>0$. 
\begin{equation}
    \label{eq:kernel}
    K_{FK}(x, x') = \frac{1}{M}\sum_{m=1}^M K_{RQ}(FK_m(x), FK_m(x'))
\end{equation}

\subsection{Diffusion Score Equilibrium}
We present our few-shot learning approach DSE shown in Figure \ref{img:meth} in this section. As elaborated in Section \ref{meth:nmg}, the weighted addition of scores $\sum_{i=1}^{N}{w_i\hat{\epsilon}_{i;\theta}(a_t, t)}$ tends towards a minimum of the weighted mixture of energy functions. The weighted scores reach an equilibrium at the minimum, or a maximum in terms of the composed probability distribution $\prod_{i=1}^{N} q_i(a_t)^{w_i}$. We use this fact to tune the composition weights $w_i$ to reduce the distance between the resulting energy minima and that of the few-shot data distribution.

Assuming $M$ motion demonstrations $D_j$ where $j=1..M$, we want to learn the optimal policy, which we evaluate using the MMD-FK distance between the data-distribution and samples from the policy. Given the limited number of demonstrations, the policy trained on the few-shot data learns a very noisy estimate of the score function. Sampling from such a policy often results in incorrect motions as the energy function gradient estimates are not accurate. \emph{Our main insight is to use gradient priors from the base set of policies to get a more accurate estimate of actual gradient towards the minimum.} Prior compositional approaches combine diffusion policies as the weighted sum of scores of the base policies, where the weights are empirically decided based on sample quality and need not necessarily sum to $1$ \cite{du2023reduce}. A parallel to our approach would be $\sum_{i=1}^{N}{w_i\hat{\epsilon}_{i;\theta}(a_t, t)}$ where $\sum_{i=1}^{N}{w_i}=1$ for $N$ base policies, which forms our baseline. We use this score estimate as a prior for our policy learned on the few-shot data $w_{comp}\hat{\epsilon}_{comp;\theta}(a_t, t) + w_{fs}\hat{\epsilon}_{fs;\theta}(a_t, t)$ where $w_{comp} + w_{fs} = 1$. This can be reformulated as  $\sum_{i=1}^{N+1}{w_i\hat{\epsilon}_{i;\theta}(a_t, t)}$ where $\sum_{i=1}^{N+1}{w_i}=1$, where the $(N+1)^{th}$ policy is trained on the few-shot demonstrations $D$. Finally, we estimate $w_i$ by minimizing MMD-FK between the few-shot demonstration data and our composed policy samples.

Estimating $w_i$ is challenging, but attempts have been made previously to estimate the sampling parameters in differentiable samplers for diffusion models~\cite{watson2022learning} with gradient based methods. These gradient based methods are computationally expensive due to multiple backward passes through the model. Instead, we utilize a non-gradient based quadratic optimizer \cite{kraft1988software} to tune our weights with the objective function of MMD-FK. Our approach is described in Algorithm \ref{algo_dse}.

\begin{algorithm}
 \caption{DSE: Compositional Weight Estimation}
 \label{algo_dse}
 \begin{algorithmic}[1]
 \renewcommand{\algorithmicrequire}{\textbf{Input:}}
 \renewcommand{\algorithmicensure}{\textbf{Output:}}
 \REQUIRE Base policies $p_i$, $i=1..N$; Demonstrations $D$ 
 \ENSURE  Compositional weights $w_i$ \\
 \textit{\textbf{Initialize}} : Train a diffusion model $p_{N+1}$ on the demonstration data $D$ \\

 \textit{\textbf{Minimize MMD-FK}}: \\
 \FOR{$l = 1$ to $OPT\_ITER$}
     \STATE \textit{\textbf{Initialize}} : $w_i$, $\sum_{i=1}^{N+1}w_i=1$ \\
     \FOR{$k = 1$ to $NUM\_SAMPLES$}
        \FOR{$j = 1$ to $NUM\_INFERENCE\_STEPS$}
            \FOR{$i = 1$ to $N+1$}
                \STATE Obtain $\hat{\epsilon}_{i;\theta}(a_t, t)$
            \ENDFOR
            \STATE $\hat{\epsilon}_{comp} =\sum_{i=1}^{N+1}{w_i\hat{\epsilon}_{i;\theta}(a_t, t)}$
        \ENDFOR
     \ENDFOR
     \STATE Calculate $MMD{\text -}FK(SAMPLES, D)$
 \ENDFOR    
 \RETURN $w_i$, $i=1..N+1$
 \end{algorithmic}
\end{algorithm}


\section{Experiments}
\label{sec:res}
\subsection{Data Generation and Model Architecture}
We use prior motions corresponding to a line, a circle and inverted pendulum along the X, Y and Z axis as base policies for most of our experiments, unless explicitly specified. We generate joint-position demonstration data using damped-least squares based differential inverse kinematics \cite{buss2004introduction} for Franka Research-3 robot 
in Mujoco \cite{todorov2012mujoco}. The priors execute these trajectories in task space with random initial end-effector orientations and positions. All our policies are trained on the smallest variant of DiT \cite{peebles2023scalable}, conditioned on the initial state of the robot in configuration space. The model $\hat{\epsilon}_\theta(a_t, o, t)$ learns to predict the noise that was added to the input $a_t$, conditioned on the diffusion time-step $t$ and the observation $o$ using AdaLN \cite{perez2018film}.

\begin{figure*}
\includegraphics[width=\linewidth, height=1.1in]{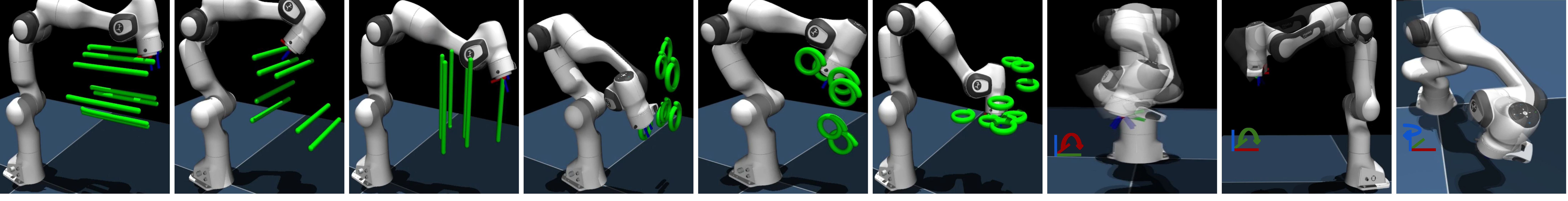}
\caption{Base policies in order: $LineX$, $LineY$, $LineY$, $CircleX$, $CircleY$, $CircleZ$, 
$OscX$, $OscY$, $OscZ$. The last three base policies $Osc$ oscillate about the specified axis with fixed end-effector position.}
\label{img:res0}
\vspace{-0.5cm}
\end{figure*}


\subsection{Baselines}
We utilize two baselines to compare against our approach. The first is the composition of diffusion policies as proposed by \citet{du2023reduce,du2020compositional}. We find optimal compositional weights for this method using the optimization procedure similar to ours. The sample size for the optimization procedure is adjusted based on the number of demonstrations in the few-shot dataset. The second is a non-compositional baseline of a diffusion model trained on the demonstration data.

\subsection{Composition and Spatial Blending}
\label{res:comp}
\begin{figure}
\minipage{0.12\textwidth}
  \includegraphics[width=\linewidth]{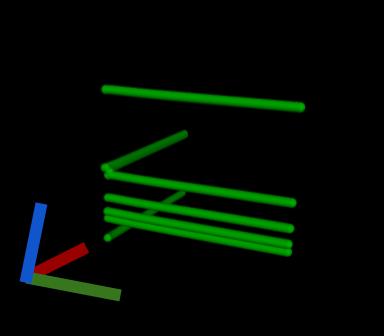}
\endminipage\hfill
\minipage{0.12\textwidth}
  \includegraphics[width=\linewidth]{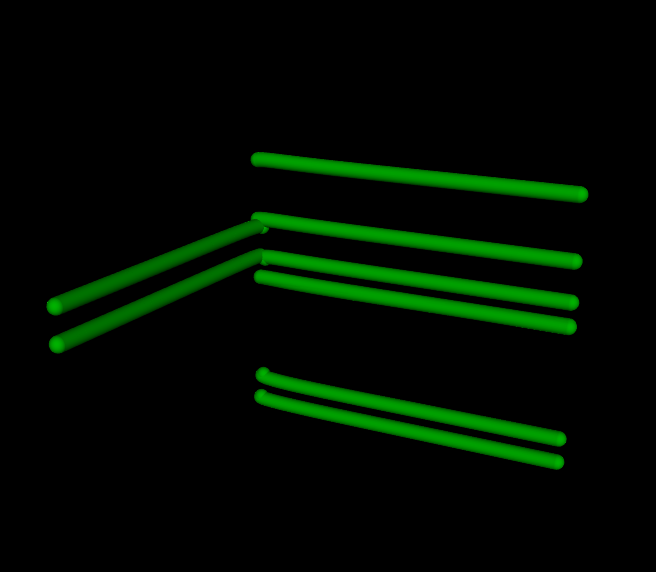}
\endminipage\hfill
\minipage{0.12\textwidth}%
  \includegraphics[width=\linewidth]{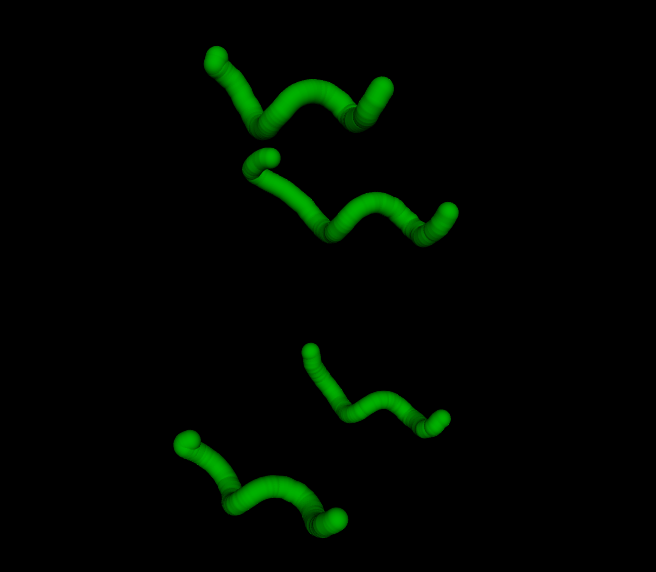}
\endminipage\hfill
\minipage{0.12\textwidth}%
  \includegraphics[width=\linewidth]{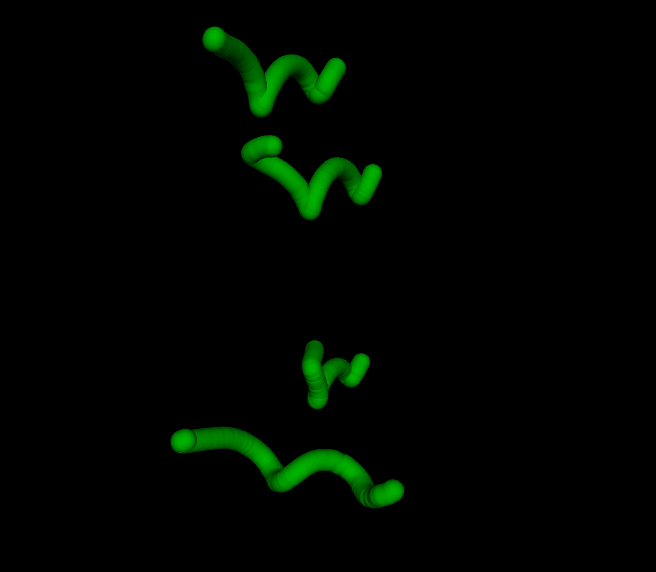}
\endminipage
\caption{This panel of figures shows A: Demo data for $+X/+Y$data. B: Demo data for $+X/-Y$data. C: Policy rollout of composition of $LineX$ and $CircleX$. D: Composition of  Line$+X/+Y$ and $CircleX$}
\vspace{-0.5cm}
\label{img:res1}
\end{figure}

We train multi-modal priors to test compositional approach's ability to sample from regions of high probability in both the distributions as shown in image A and B of Figure \ref{img:res1}. We train policy A to reach towards the $+X$ or $+Y$ direction and policy B to reach towards the $+X$ or the $-Y$ direction. We expect the composed policy C with $w_1=w_2=0.5$ to sample from the modes of reaching towards the $+X$ direction as the $+X$ behavior exists in both Policy A and B. 
We see exactly this behavior as the MMD-FK between Policy A and policy C is $0.58$, between Policy B and Policy C is $0.27$ and Policy C and a $+X$ direction policy is $0.11$. Lower values of MMD-FK indicates lower errors or higher match between the two trajectory distributions. 
Composing policies to sample from the common regions of high probability was also shown for the reach and obstacle avoidance task by Urain et al.~\cite{urain2023composable}. However, their work used hand crafted potential functions to compose these distributions~\cite{urain2023composable}. We also showcase spatial blending where we compose a policy $CircleX$ and policy $LineX$ to create a spiral, as shown in image C Figure~\ref{img:res1}. The MMD-FK metrics obtained for both the cases are provided in Table~\ref{tab:1}. Finally, we showcase the result of composing the multi-modal policy Line$+X/+Y$ and policy $CircleX$ in image D in Figure \ref{img:res1}. The composed policy is more dominant along the $+Y$ direction due to the directional similarity of motions.

\begin{table}[h]
    \caption{MMD-FK values between samples from the composed and the base policy distributions. The compositional weights are taken to be $w_1=w_2=0.5$ for both cases. Self-Comparison implies that the MMD-FK is calculated between demonstration data and rollouts for the same policy.}
    \label{tab:1}
    \centering
    \begin{tabular}{|c|c|c|c|} \hline
                    & +X & CircleX        \\ \hline
        Spiral Vanilla Composition & 0.92  & 0.87          \\ \hline
        Self-Comparison  & 0.03  & 0.01          \\ \cline{1-3}
    \end{tabular}
\end{table}

\subsection{Few-shot learning}
\label{res:fsl}

\begin{figure}
\minipage{0.12\textwidth}
  \includegraphics[width=\linewidth]{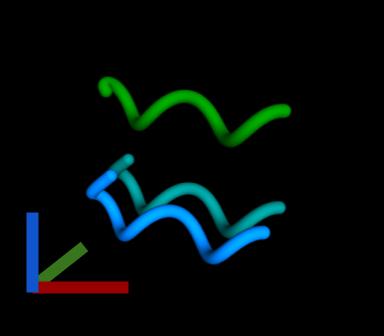}
\endminipage\hfill
\minipage{0.12\textwidth}
  \includegraphics[width=\linewidth]{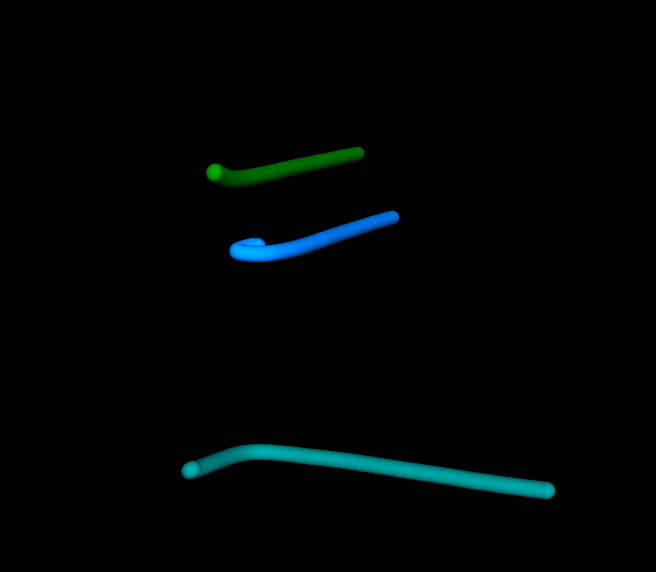}
\endminipage\hfill
\minipage{0.12\textwidth}%
  \includegraphics[width=\linewidth]{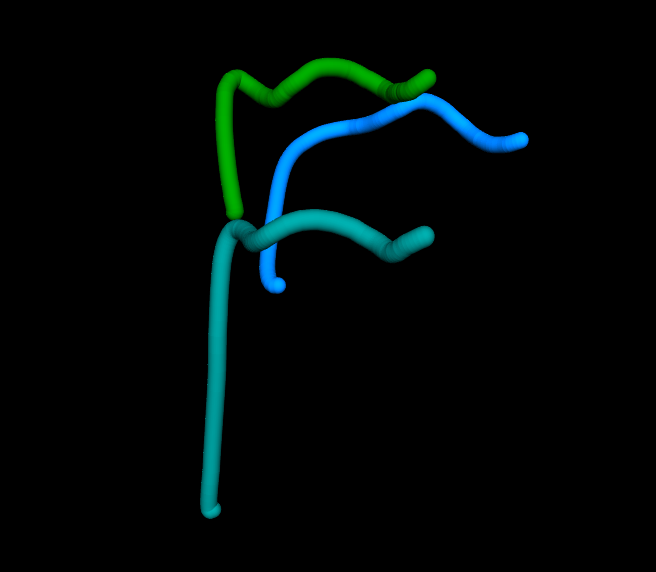}
\endminipage\hfill
\minipage{0.12\textwidth}%
  \includegraphics[width=\linewidth]{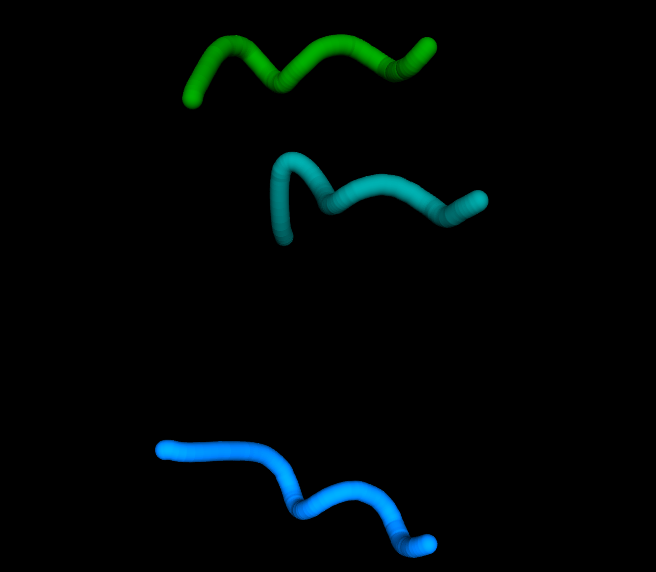}
\endminipage
\caption{This panel of figures shows A: EEF Demo data for spiral trajectory. B: Policy rollout of vanilla composition C: Policy rollout of the fine-tuned policy trained on 15 demos D: Policy rollout of DSE trained on 15 demos.}
\label{img:res2}
\vspace{-0.5cm}
\end{figure}

We present our few-shot learning results in this section. A core element of our approach is the optimization procedure to evaluate the compositional weights. For all the experiments, we run the optimization procedure 4 times, where it is initialized with the normalized MMD-FK values between the prior motion datasets and the novel demonstration dataset, and three random initial values that sum to 1. We found that the optimization was also able to recover the base policies from corresponding demonstration data collected on the real robot. We compare DSE against our baselines for 5 novel trajectories not seen by the robot, three in a simulated setting, and two collected on the real robot. We report MMD-FK values with the reference trajectory distribution wherever available, evaluated over $50$ samples. We also
report the mean squared error values with the trajectories
collected on the real robot for all the policies. Table \ref{res:fse} shows the results for the simulated experiments. DSE consistently achieves a lower or comparable MMD-FK score than both the baselines on all the three tasks, for $5$, $15$ and $40$ demonstrations. While we
visually represent the end effector trajectories in the paper, our
method optimizes the compositional weights for all the links
of the robot. The video rollouts of the composed trajectories can also be viewed on our webpage\footnote{\url{https://sites.google.com/asu.edu/comp-fsl}}. We analyze each experiment closely below. 
\begin{itemize}[leftmargin=*, nolistsep]
    \item \textbf{SpiralX}: We generate a spiral trajectory along the X-axis as the target policy. In this experiment, along the lines of Section \ref{res:comp}, we consider only Line$+X/+Y$ and $CircleX$ as our prior policies. The vanilla composition method clearly struggles in this case due to the prior policy being multi-modal, as explained in Section \ref{res:comp}. DSE performs the best of the three approaches compared, as shown in Figure \ref{img:res2}.
    \item \textbf{Step}: We generate a step trajectory in the XZ plane. We observe that DSE policy performs surprisingly well with just 5 demonstrations, largely due to the base policy gradient priors, while the fine-tuned policy does not perform well. As the number number of demonstrations is increased, the fine-tuned policy catches up to DSE in terms of MMD-FK.    
    \item \textbf{OscX+LineXZ}: We create a difficult target distribution for the final case in the simulated setting. The robot end effector moves along a line while the robot body is oscillating about the X axis. We observe that the fine-tuned policy performance gets better with increasing number of demonstrations while compositional weight optimizer struggles due to the small oscillatory movements in the target.
\end{itemize}

\begin{table}[h]
\centering
\begin{tabular}{|c|c|c|c|c|}
    \hline
      & Number   & Vanilla  & Fine-tuned &Diffusion \\
                      & of  &Composition& Policy                        & Score \\
    Trajectories & demos  &&                              & Equilibrium\\\hline
   
    \multirow{3}{4em}{Spiral X}  
    & 5 & \textbf{0.58} & 0.64 &\textbf{0.51}  \\ \cline{2-5}
    & 15& 0.58 & 0.26 & \textbf{0.09} \\ \cline{2-5}
    &  40& 0.58 &0.15  & \textbf{0.09} \\ \hline
    \multirow{3}{4em}{Step}  
    &5& 0.79 & 0.50 & \textbf{0.25} \\ \cline{2-5}
    &15 & \textbf{0.18} & 0.27 & 0.20 \\ \cline{2-5}
    &40 &\textbf{0.15} & \textbf{0.17} &\textbf{0.12}  \\ \hline
    \multirow{3}{4em}{OSC X  + Line XZ}
    & 5& 0.75 & 0.57 & \textbf{0.32}\\ \cline{2-5}
    & 15& 0.30 & 0.25 & \textbf{0.06} \\ \cline{2-5}
    & 40& 0.37 & \textbf{0.14} & \textbf{0.12 }\\ \hline
\end{tabular}
\caption{MMD-FK scores for $50$ rollouts across skills and  demonstrations counts. DSE out-performs both our baselines consistently.}
\label{res:fse}
\end{table}

\begin{figure}
    \centering
    \includegraphics[width=1\linewidth]{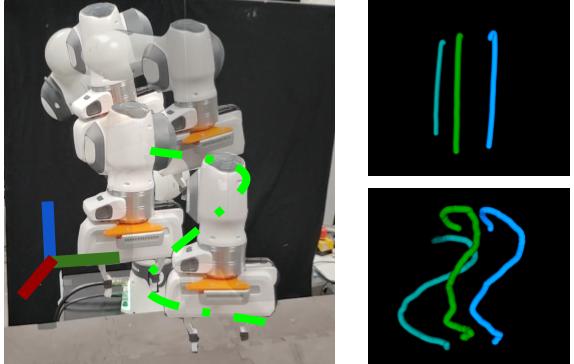}
    \caption{This panel of figures shows Left: Overlay of real robot demonstration collection; Top-right: Policy rollout of vanilla composition with 15 demos; Bottom-right: Policy rollout of DSE trained on 5 demos.}
    \label{img:res3}
    \vspace{-0.5cm}
\end{figure}

For our real world experiment, we collected $15$ demonstrations resembling an $S$ along the x-axis and Spring motion along x-axis. The MMD-FK results are shown in Table \ref{res:realrobot} and visually represented in Figure \ref{img:res3}. We also show the mean squared error(MSE) between the collected demonstrations on the real robot and the rolled out trajectory from the corresponding initial states. DSE also achieved lower MSE with the collected demonstrations than the baselines, confirming the utility of our metric MMD-FK for evaluating compositional weights.

\begin{table}[h]
\centering
\begin{tabular}{|c|c|c|c|c|}
    \hline
      & Number   & Vanilla  & Fine-tuned &Diffusion \\
                      & of  &Composition& Policy                        & Score \\
    Trajectories & demos  &&                              & Equilibrium\\\hline

    \multirow{2}{4em}{S Motion}  
    & 5 & \textbf{0.50} / 0.0076 & 0.69 / 0.0034 &\textbf{0.56} / \textbf{0.0019 } \\ \cline{2-5}
    & 15& 1.70 / 0.0148 & 0.69 / 0.0023 & \textbf{0.34} / \textbf{0.0015} \\ \hline
    
    \multirow{2}{4em}{Spring Motion}  
    & 5 & 1.65 / 0.016 & 4.28 / 0.0037 &\textbf{0.37} / \textbf{0.0024 } \\ \cline{2-5}
    & 15& 0.91 / 0.0110 & 5.10 / 0.0022 & \textbf{0.47} / \textbf{0.0013} \\ \hline

\end{tabular}
\caption{Robot experiment results where we collected $15$ demonstrations on Franka FR3 to train our policies. DSE achieves lower MMD-FK/MSE values exhibiting robustness to noise when learning.}
\label{res:realrobot}
\end{table}

\vspace{-0.5cm}
\section{Discussion and Limitations}
As the number of training demonstrations are increased, the weight assigned by our approach DSE to the fine-tuned model increases. This is expected as if we have more demonstrations our model picks the true data distribution rather than the compositions over the base policies. However, as we observe more data vanilla composition models also perform better as they get a better estimate of the trajectory distribution. 

Our priors are not orthogonal and can be chosen with a lot of freedom. The DSE method also works with multi-modal prior policies, as showcased in Section~\ref{res:comp}. This is unlike policy composition using multiplicative Gaussian policies~\cite{peng2019mcp} which cannot handle multi-modality. Moreover, Gaussian Mixture Models face the challenge of exploding number of modes as the number of prior policies increase, further highlighting the efficiency of DSE. Our results can also improve with more priors however this would lead to increased compute time to find optimal weights.

The metric MMD-FK is created to evaluate the distance between two trajectory distributions for the whole body of the robot. However, we can re-weight the links to pay more emphasis on the end-effector or any other link as required. 
Empirically we see that our fine-tuned policy forms an upper-bound for the MMD-FK score obtained for DSE.

We do want to acknowledge that these compositions are in the state space of the robot rather than in the raw observation space such as the visual observations of the robot.

\section{Conclusion}
\label{sec:conclusion}
We present a novel compositional approach to few-shot learning called Diffusion Score Equilibrium (DSE) based on equilibrium of scores predicted by diffusion models. Our approach composes a policy trained on the target demonstrations with a set of base policy priors and infers the compositional weights by minimizing a measure of distance between the resulting composed distribution and the demonstration data distribution. Empirically, we observed that DSE will perform better than a policy simply trained on the data irrespective of the number of provided demonstrations on average by $30\% - 50\%$, while outperforming it by significant margins in the few-shot regime. We also propose a novel metric MMD-FK to measure the distance between two movement trajectory distributions for the whole body of the robot.

\bibliography{bib/IEEEabrv.bib,bib/references.bib}{}
\bibliographystyle{IEEEtranN}
\end{document}